\documentclass[10pt,twocolumn,letterpaper]{article}

\usepackage[pagenumbers]{cvpr} % To force page numbers, e.g. for an arXiv version

\definecolor{cvprblue}{rgb}{0.21,0.49,0.74}
\usepackage[pagebackref,breaklinks,colorlinks,allcolors=cvprblue]{hyperref}

\title{Emotion-Director: Bridging Affective Shortcut in \\ Emotion-Oriented Image Generation}

\author{Guoli Jia$^{1}$* \hspace{0.5cm} 
Junyao Hu$^{2}$* \hspace{0.5cm} 
Xinwei Long$^1$ \hspace{0.5cm} 
Kai Tian$^{1,3}$ \hspace{0.5cm} 
Kaiyan Zhang$^{1,3}$ \hspace{0.5cm}
KaiKai Zhao$^{1,4}$ \hspace{0.5cm}
\\
Ning Ding$^{1}$  \hspace{0.5cm} 
Bowen Zhou$^{1,5 \dagger}$ 
\vspace{0.5em}
\\
$^1$ Tsinghua University \hspace{0.6cm} $^2$ The Hong Kong Polytechnic University  \hspace{0.6cm} $^3$ Frontis.AI 
\\
$^4$ China Unicom \hspace{0.6cm}
$^5$ Shanghai Artificial Intelligence Lab
\vspace{0.5em}
\\
{\small $^{*}$Equal contribution.}  \hspace{0.8cm} {\small $^{\dagger}$Corresponding author.}
}

\usepackage{hyperref}       % hyperlinks
\usepackage{url}            % simple URL typesetting
\usepackage{booktabs}       % professional-quality tables
\usepackage{amsfonts}       % blackboard math symbols
\usepackage{nicefrac}       % compact symbols for 1/2, etc.
\usepackage{microtype}      % microtypography
\usepackage{xcolor}         % colors
\usepackage{amsmath}
\usepackage[capitalize]{cleveref}
\usepackage{enumitem}
\usepackage{graphicx}
\usepackage{caption}
\usepackage{adjustbox}
\usepackage{array}
\usepackage{hyperref}
\usepackage{colortbl}

\usepackage{newfloat}
\usepackage{listings}

\usepackage{multirow}
\usepackage{caption}
\usepackage{array}
\usepackage{amsmath,amssymb}
\usepackage{dsfont}
\usepackage{colortbl}
\usepackage{xcolor} 
\usepackage{booktabs}
\usepackage{float}

\begin{document}

\twocolumn[{
\maketitle
\begin{center}
  \centering
  \includegraphics[width=1.0\linewidth]{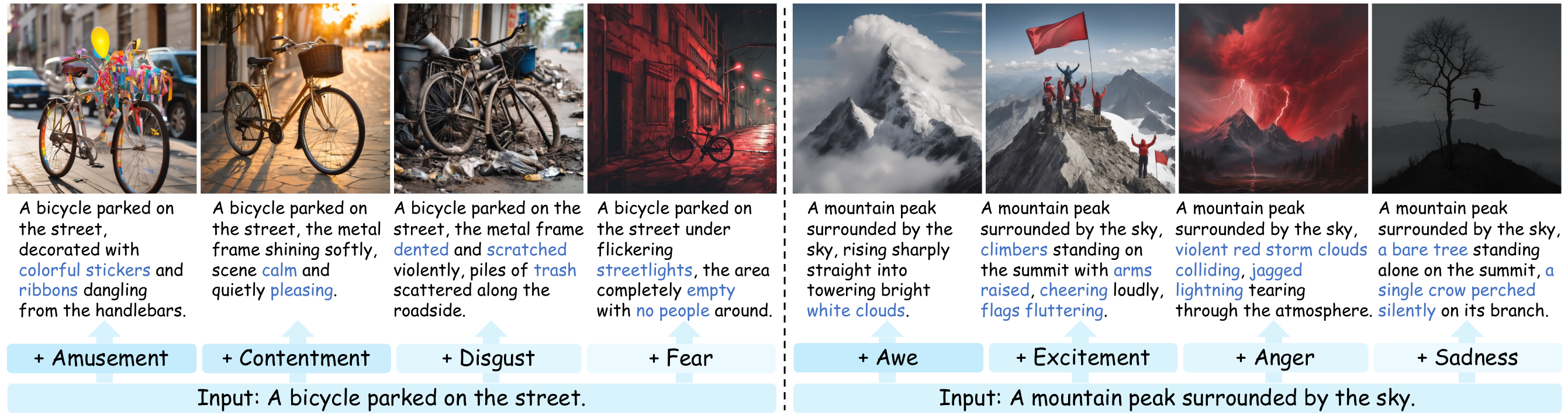}
  \vspace{-10pt}
   \captionof{figure}{
    Two cases of emotion-oriented image generation. Given a textual prompt and an intended emotion, our proposed Emotion-Director rewrites the prompt based on the specified emotion, then generates \textit{emotion-highlighted} images.
   }
   % \vspace{-5pt}
   \label{fig:teaser}
\end{center}
}]

\maketitle
\begin{abstract}
Image generation based on diffusion models has demonstrated impressive capability, motivating exploration into diverse and specialized applications.
Owing to the importance of emotion in advertising, emotion-oriented image generation has attracted increasing attention.
However, current emotion-oriented methods suffer from an affective shortcut, where emotions are approximated to semantics.
As evidenced by two decades of research, emotion is not equivalent to semantics.
To this end, we propose Emotion-Director, a cross-modal collaboration framework consisting of two modules.
First, we propose a cross-Modal Collaborative diffusion model, abbreviated as MC-Diffusion.
MC-Diffusion integrates visual prompts with textual prompts for guidance, enabling the generation of emotion-oriented images beyond semantics.
%
% Further, we design an improved DPO algorithm to enhance the model's sensitivity to different emotions under the same semantics.
Further, we improve the DPO optimization by a negative visual prompt, enhancing the model's sensitivity to different emotions under the same semantics.
Second, we propose MC-Agent, a cross-Modal Collaborative Agent system that rewrites textual prompts to express the intended emotions.
To avoid template-like rewrites, MC-Agent employs multi-agents to simulate human subjectivity toward emotions, and adopts a chain-of-concept workflow that improves the visual expressiveness of the rewritten prompts.
Extensive qualitative and quantitative experiments demonstrate the superiority of Emotion-Director in emotion-oriented image generation.
\end{abstract}
    
\section{Introduction}
\label{sec:intro}

Image generation~\cite{ho2020denoising, peebles2023scalable} has achieved unprecedented success~\cite{rombach2022high, podellsdxl}.
To facilitate its real-world applications, such as advertisement intelligence~\cite{du2024towards, holbrook1984role} and smart tourism~\cite{zhang2024linking, zhang2023influence}, emotion-oriented image generation has drawn increasing attention~\cite{yang2024emogen, yang2025emoedit, mao2025emoagent}.
Despite the remarkable progress achieved in affective image generation~\cite{zhang2025affective, zhu2025uniemo, yuan2025coemogen}, they still face the challenge of \textit{\textbf{affective shortcut}}.

\begin{figure*}[t!]
  \centering
  \includegraphics[width=0.96\linewidth]{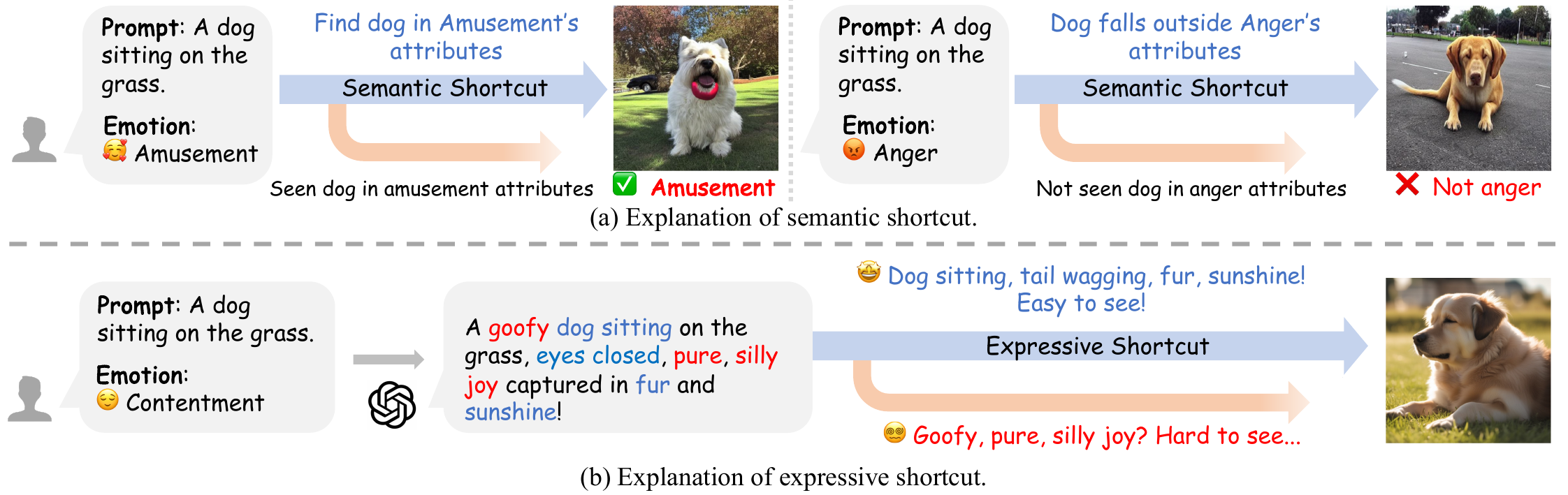}
   \vspace{-10pt}
   \caption{
   Illustration of affective shortcut, which consists of \textit{semantic shortcut} and \textit{expressive shortcut}.
   (a) Semantic shortcut: Current methods tightly couple emotions with semantic attributes. As a result, they fail to generate emotion-faithful images when the prompt’s semantics fall outside the emotion-specific attributes.
   Using above prompt as an example, we generate 10 images for each emotion, only 38.75\% simultaneously contain a dog and convey the intended emotion.
   (b) Expressive shortcut: Directly using LLMs to rewrite prompts often leads to template-like outputs that cannot be visually expressed.
   Using the above prompt as an example, we employ Qwen2.5 to generate 10 rewritten prompts for each emotion. 
   For both the original and rewritten prompts, we use SDXL to generate 4 images and compute the average text–image consistency. 
   The CLIP Score decreases from 35 to 31.
   }
   \vspace{-15pt}
   \label{fig:motivation}
\end{figure*}

The affective shortcut consists of two specific challenges, as illustrated in Fig.~\ref{fig:motivation}.
1) \textit{\textbf{Semantic shortcut}}: Current methods learn emotion representations by aligning emotion with semantic attributes~\cite{yang2024emogen, zhu2025uniemo}.
However, as evidenced by two decades of research~\cite{zhao2021affective}, \textit{\textbf{emotion is not equivalent to semantics}}, the same semantics can evoke entirely different emotions (please see Fig.~\ref{fig:comparison-2}).
In practice, since ``dog'' is rarely associated with anger, the model fails to generate ``angry'' images that include a dog.
This indicates that relying on semantic attributes is essentially a shortcut rather than a proper modeling of emotion.
2) \textit{\textbf{Expressive shortcut}}: A straightforward way to alleviate the semantic shortcut is to utilize text guidance, where a large language model (LLM)~\cite{kong2024hunyuanvideo, qwen2.5} rewrites texts to enrich the semantics required for expressing the intended emotion.
However, we observe that the rewritten texts are often template-like, expressing emotions by adding descriptions that cannot be visually depicted, \textit{e.g., ``goofy'', ``silly joy''}, resulting in a noticeable drop in CLIP Score~\cite{radford2021learning}.
To this end, we propose a method that bridges the affective shortcut, anthropomorphically named Emotion-Director.
The core insight of Emotion-Director is \textit{\textbf{cross-modal collaboration}}, which consists of two modules.
First, \textit{\textbf{we propose a cross-Modal Collaborative diffusion model, abbreviated as MC-Diffusion}}.
MC-Diffusion integrates visual prompts into guidance conditions, forming multi-modal prompts to enable the model to express emotions beyond textual semantics.
To adapt visual prompts to diverse semantics, MC-Diffusion constructs a prompt bank that stores visual tokens capable of collaborating with various textual prompts.
Furthermore, we adopt DPO for training, and improve the loss with a negative visual prompt term to enhance model's sensitivity to different emotions under the same textual semantics.
%
% The DPO loss also alleviates the ambiguity in emotion~\cite{jia2022s} by learning preference relations of emotion, rather than directly fitting ambiguous boundaries between emotions as in previous SFT methods~\cite{yang2024emogen, lin2024make}.
%
Second, \textit{\textbf{we develop a cross-Modal Collaborative Agent system, abbreviated as MC-Agent}}.
%
% MC-Agent consists of four procedures, including visual concept extraction, emotion attribution, textual prompt rewriting, and checking.
%
The multiple agents simulate users' subjectivity to emotions, thereby alleviating the template-like bias in text rewriting.
Moreover, the visual concepts, \textit{such as humans, animals, and landscapes}, flow among multiple agents, serving as a bridge between textual descriptions and visual expressiveness.
The ``chain-of-concept'' workflow supports MC-Agent to generate emotion-highlighted and visually expressible textual prompts.
Our contributions are summarized as:
\begin{itemize}[leftmargin=*]
    \item We explore the affective shortcut in current emotion-aware methods, and propose a cross-modal collaboration framework named Emotion-Director to overcome this limitation.
    \item We design MC-Diffusion and MC-Agent, with technical contributions including multi-modal prompts, improved DPO loss, and chain-of-concept workflow for emotion-oriented image generation.
    \item The extensive qualitative and quantitative experiments consistently demonstrate the superiority of Emotion-Director.
\end{itemize}

\section{Related Work}
\label{sec:related}

\subsection{Image Generation}
Traditional image generation is primarily built on GANs~\cite{brocklarge, goodfellow2014generative, karras2019style} and VAEs~\cite{kingma2013auto, van2017neural}, has rapidly advanced with the recent breakthroughs in diffusion models~\cite{podellsdxl, rombach2022high}.
Currently, leading image generation models have achieved remarkable progress in controllability~\cite{zhang2023adding, ruiz2023dreambooth}, efficiency~\cite{11002717, song2023consistency}, and reasoning capability~\cite{xue2025dancegrpo, liu2025flow}.
Diffusion models have significantly improved the quality and diversity of image generation~\cite{rombach2022high, ho2020denoising}. 
To further enhance controllability, text guidance~\cite{podellsdxl, esser2024scaling} and customized fine-tuning~\cite{ruiz2023dreambooth, chen2024anydoor, wang2025ps} are proposed for training.
However, the iterative denoising process introduces high computational costs and low sampling efficiency~\cite{songdenoising}. 
To address this issue, acceleration strategies such as consistency models~\cite{song2023consistency, luo2023latent}, ODE-based sampling~\cite{liupseudo, lu2022dpm}, and distillation~\cite{salimansprogressive} are proposed for fast sampling.
Recently, inspired by the success of reinforcement learning in large language models (LLMs)~\cite{zhang2025survey, guo2025deepseek}, post-training is applied to diffusion~\cite{xue2025dancegrpo, liu2025flow} and auto-regressive~\cite{jiang2025t2i} models, substantially improving their reasoning capabilities over attributes, counting, spatial relationships.
\textit{\textbf{In this paper, we explore emotion-oriented image generation to advance the powerful foundational capability towards real-world applications}}.

\subsection{Affective Computing}
Affective computing has been explored for over two decades~\cite{zhao2021affective}, with early research primarily focusing on emotion understanding, \textit{e.g.}, \textit{recognition}~\cite{jia2022s, feng2023probing}, \textit{grounding}~\cite{yang2018weakly, zhang2022temporal}.
With the rapid advancement of generative foundation models, emotion-oriented generative models exhibit broader potential in real-world applications such as tourism~\cite{zhang2023influence, zhang2024linking} and advertising~\cite{du2024towards}.
Most existing emotion-oriented image generation methods map emotions to semantic attributes or prototypes~\cite{yang2024emogen, zhu2025uniemo, yuan2025coemogen}.
However, inspired by the experience in previous emotion recognition research~\cite{zhao2021affective, zhao2014exploring}, this semantic alignment suffers from semantic shortcuts, and their lack of controllability significantly limits user's practical usability.
Notably, unlike these methods, emotion-oriented image editing methods~\cite{yang2025emoedit, lin2024make, weng2023affective, zhangtowards} attempt to convey emotions by adjusting visual details, such as color and texture, in the user-provided source image. 
However, unlike subject-driven editing~\cite{hertz2023prompt}, the abstract of emotion~\cite{yang2018weakly} makes it difficult to localize the regions that should be modified or preserved.
Therefore, the edited image exhibits nearly the same semantics as the source image, and only editing visual detail is usually insufficient to convey the intended emotion.
\textit{\textbf{To this end, we design a cross-modal collaboration framework to enable emotion-highlighted image synthesis}}.

\subsection{Direct Preference Optimization in Diffusion Models}
Direct Preference Optimization (DPO)~\cite{ouyang2022training, rafailov2023direct} has demonstrated remarkable advantages in human feedback alignment, which trains models to directly learn the preference relation from paired samples.
Recently, DPO has been successfully applied to align diffusion models~\cite{yang2024using, wu2024multimodal}.
As a pioneering approach, Diffusion-DPO~\cite{wallace2024diffusion} trains diffusion models on the Pick-a-Pic dataset with preference annotation, showing a noticeable improvement in image quality and text fidelity.
Considering the importance of preference pairs in DPO, D3PO~\cite{yang2024using} employs human evaluators to select preference images, and VisionPrefer~\cite{wu2024multimodal} validates the preference annotation ability of multi-modal large language models (MLLM). 
Further, negative preference optimization (NPO)~\cite{wangdiffusion} incorporates negative sample pairs into the DPO inspired by classifier-free T2I generation.
Recently, step-by-step preference optimization (SPO)~\cite{liang2024step} trains a step-aware preference model with noisy latents, significantly improving the aesthetic quality of diffusion models.
Considering the ambiguous boundaries between emotions~\cite{jia2022s}, we employ DPO to optimize our MC-Diffusion, which utilizes contrastive preference pairs to improve its sensitivity to emotion, instead of directly learning ambiguous boundaries.

\begin{figure*}[!t]
  \centering
  \includegraphics[width=0.96\linewidth]{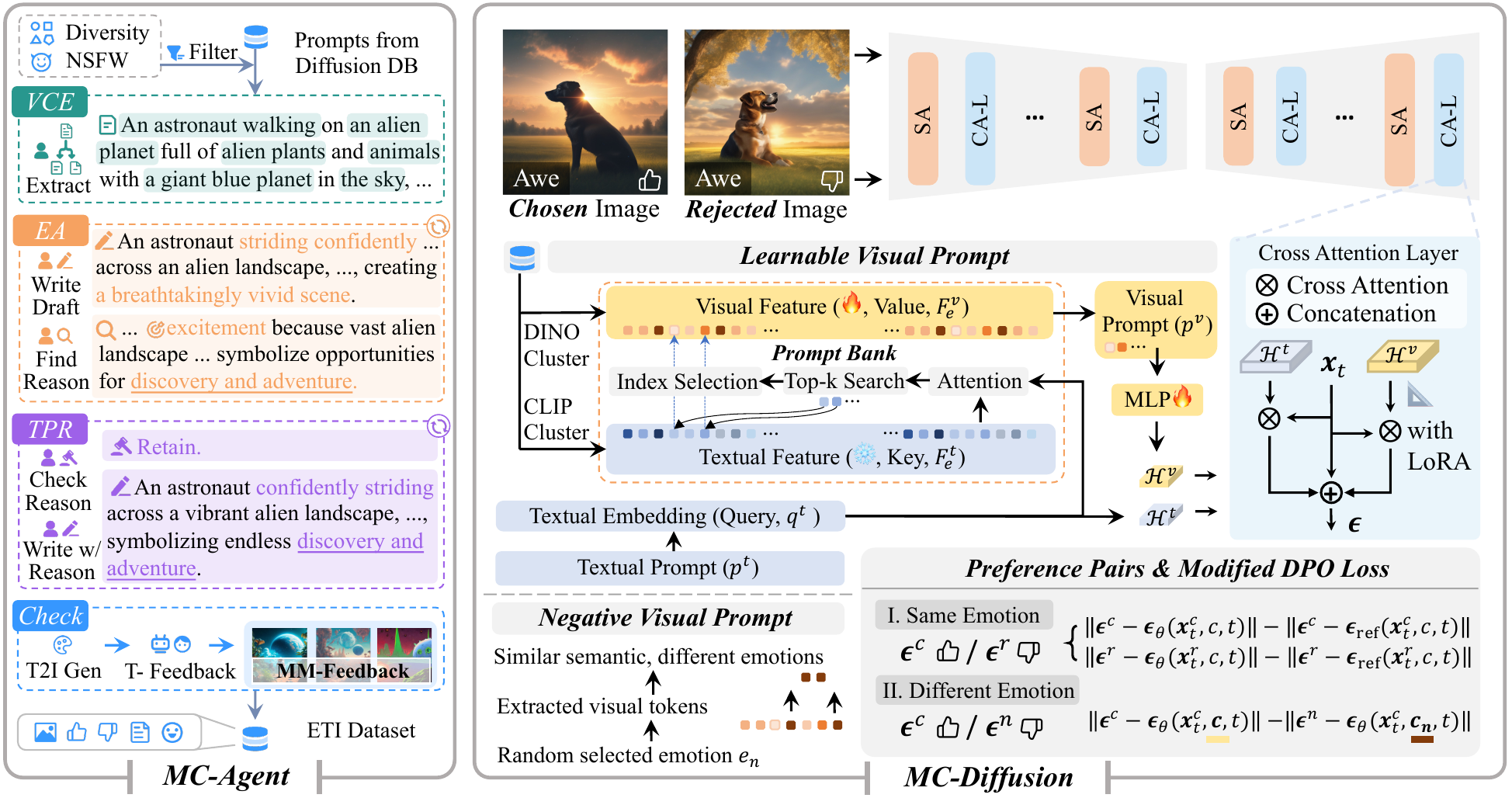}
   \vspace{-6pt}
   \caption{Pipeline of Emotion-Director. 
    % \textit{MMP-Agent} (left) implicitly integrates textual and visual information through four steps to construct a high-quality, emotionally diverse dataset, named ETI.
    VCE, EA, TPR, and Check represent the four procedures of the MC-Agent: visual concept extraction, emotion attribution, textual prompt revision, and checking.
    %
    % T-Feedback and MM-Feedback represent the textual and multi-modal feedback respectively.
    % 
    % \textit{MMP-Diffusion} (right) effectively achieves emotion-oriented text-to-image generation by simultaneously learning textural and visual prompts and modifying the original DPO loss function.}
    SA and CA refer to self-attention and cross-attention in the diffusion model, respectively.
    CA-L indicates that LoRA is applied to the cross-attention.}
   \vspace{-15pt}
   \label{fig:pipeline}
\end{figure*}

\section{Methodology}
\label{methodology}

We propose Emotion-Director, which bridges the affective shortcut in emotion-oriented image generation via a cross-modal collaboration framework.
As shown in Fig.~\ref{fig:pipeline}, Emotion-Director consists of MC-Diffusion and MC-Agent.
The input to Emotion-Director includes the user-specified textual prompt $p$ and an emotion $e$, where $e$ is one of the emotion classes defined in the emotion model $E$.
Here, we adopt Mikels emotion model~\cite{mikels2005emotional}, which is the most commonly used model in visual emotion analysis.
MC-Agent rewrites $p$ according to $e$, and its output is defined as $p^t$.
Subsequently, $p^t$ and $e$ are fed into MC-Diffusion.
MC-Diffusion obtains the visual prompt $p^v$ based on the inputs, and then uses both $p^t$ and $p^v$ as guidance to generate emotion-highlighted images.
\subsection{MC-Diffusion}
MC-Diffusion utilizes multi-modal prompts and the improved DPO loss to mitigate semantic shortcut. 
Specifically, MC-Diffusion integrates the visual prompt as guidance to generate emotion-oriented images beyond semantics.
Furthermore, MC-Diffusion employs an improved DPO loss to enhance the model's emotional sensitivity under the same semantics.
%

%%% 写作思路： visual tokens怎么初始化 -> visual tokens怎么获取 -> 怎么同时输入MM prompts -> DPO怎么优化

%
\noindent \textbf{Semantics-Adaptive Multi-modal Prompts}. The textual prompt $p^t$ is rewritten by MC-Agent, as illustrated in Sec.~\ref{sec:MC-Agent}.
To enable the visual prompt $p^v$ to be adapted to diverse semantics of textual prompts, we maintain a prompt bank $\mathcal{P}^{V}$, and retrieve the suitable visual prompt according to $p^t$ and $e$.
$\mathcal{P}^{V} \in \mathbb{R}^{|E| \times L}$ is initialized with features extracted from the image-text emotion dataset, where $|E|$ denotes the number of emotions in $E$, $L$ is the number of candidate prompts for each emotion.
For each image-text pair, we extract the textual feature $f^t \in \mathbb{R}^{1 \times d^t}$ using text encoder in CLIP~\cite{radford2021learning}, and the visual feature $f^v \in \mathbb{R}^{1 \times d^v}$ using DINOv3~\cite{simeoni2025dinov3}, where $d^t$ and $d^v$ denote the feature dimensions.
Next, for each emotion $e$, we apply the K-means algorithm to divide the set of $f^t$ into $L$ clusters, denoted as $F^t_e \in \mathbb{R}^{L \times d^t}$.
To support retrieval, the closest feature to the cluster center is used as the new center to prevent modifying the clip-extracted features.
Then, the clustering labels of $f^v$ are set the same as those of $f^t$.
By computing $L$ cluster centers of $f^v$, we obtain $F^v_e \in \mathbb{R}^{L \times d^v}$.
The visual prompt bank for emotion $e$ is defined as $\mathcal{P}^{V}_{e} = \{F^t_e, F^v_e\}$.
Given $p^t$ and $e$, we obtain the textual embedding $q^t \in \mathbb{R}^{1 \times d^t}$ by feeding $p^t$ into the CLIP text encoder.
Then, we retrieve the visual prompt $p^v \in \mathbb{R}^{\ell \times d^v}$ from $\mathcal{P}^{V}_{e}$ as follows:
\begin{equation}
    p^v = F^v_e[\mathcal{I}^\ell], \quad \mathcal{I}^\ell = Topk \, (q^t \cdot ({F^t_e})^{T}), 
\end{equation}
where $\mathcal{I}^\ell$ denotes the indices of $F^t_e$ computed by CLIP that have the highest similarity with $q^t$, aiming to retrieve the most relevant images.
$\ell$ is the length of the visual prompt, and we set it the same as the textual embeddings.
Subsequently, the visual prompt $p^v$ is indexed from $F^v_e$.
Next, the multi-modal prompts guide the generation by cross-attention.
Specifically, to preserve the powerful T2I capability in pre-training, we retain the standard cross-attention to incorporate the textual prompt.
To integrate the visual prompt, we apply low rank adaptation (LoRA)~\cite{hulora} to the cross-attention layer.
Let $X \in \mathbb{R}^{b \times \ell_h \times d^{h}}$ denote the U-Net hidden states, $b$ means the batch size, $\ell_h$ and $d^h$ are the length and dimension of hidden states.
We compute two sets of projections as follows:
\begin{equation}
\begin{alignedat}{3}
Q^{t} =& W_QX,& K^{t} =& W_K\mathcal{H}^{t},& V^{t} =& W_V\mathcal{H}^{t},   \\
Q^{v} =& A_Q B_Q X,& K^{v} =& A_K B_K \mathcal{H}^{v} ,&V^{v} =& A_V B_V \mathcal{H}^{v}. \\
\end{alignedat}
\end{equation}
Here, $\mathcal{H}^* \in \mathbb{R}^{b \times \ell \times d^h}$ are the embeddings extracted from a batch of $p^t$ and $p^v$.
$W_*$ are the standard attention projection matrices, $A_* \in \mathbb{R}^{d^h \times r}$ and $B_* \in \mathbb{R}^{r \times d^h}$ are LoRA parameters.
$r$ is the rank of LoRA.
The cross-attention of two branches is calculated independently.
Then, the final output $\widetilde{\mathcal{X}}$ of the cross-attention module is obtained by linearly combining the two branches as follows:
{
    \small
    \begin{equation}
    \widetilde{\mathcal{X}} = \mathrm{CrossAttn}(Q^{t}, K^{t}, V^{t}) + \mathrm{CrossAttn}(Q^{v}, K^{v}, V^{v}).
    \end{equation}
}
\noindent \textbf{Emotion-Sensitive DPO Loss}. In addition to incorporating $p^v$ to address the semantic shortcut, we improve the direct preference optimization (DPO) algorithm to enhance the model’s sensitivity to different emotions under the same semantics.
Given preference data $\{c, x^c, x^r\}$, typical DPO loss in the diffusion model~\cite{wallace2024diffusion} derives a differentiable objective based on the evidence lower bound, encouraging the model to align with the chosen image.
It approximates the sampling in diffusion-specific likelihoods as:
{
    \small
    \begin{equation}
    \begin{aligned}
        \mathcal{L}(\theta) &= -\mathbb{E}_{\boldsymbol{x}_t^c,\boldsymbol{x}_t^r,c,t}[\log \sigma (- \beta T \omega ( \\
        & (\lVert \boldsymbol{\epsilon}^c - \boldsymbol{\epsilon}_\theta(\boldsymbol{x}_t^c,c,t) \rVert_2^2 - \lVert \boldsymbol{\epsilon}^c - \boldsymbol{\epsilon}_\mathrm{ref}(\boldsymbol{x}_t^c, c,t) \rVert_2^2) \\
         - &(\lVert \boldsymbol{\epsilon}^r - \boldsymbol{\epsilon}_\theta(\boldsymbol{x}_t^r,c,t) \rVert_2^2 - \lVert \boldsymbol{\epsilon}^r - \boldsymbol{\epsilon}_\mathrm{ref}(\boldsymbol{x}_t^r, c,t) \rVert_2^2) ) ) ],
    \end{aligned}
    \end{equation}
}%
where $\theta$ is the parameters of diffusion model.
$c$ denotes the guidance condition, $x^c_t$ and $x^r_t$ are the latents of the chosen and rejected images at timestep $t$, in total of $T$ timesteps.
$\boldsymbol{\epsilon}^c$ and $\boldsymbol{\epsilon}^r$ are the added noises.
$\boldsymbol{\epsilon}_\theta$ and $\boldsymbol{\epsilon}_\mathrm{ref}$ are the predicted noises of the optimized and reference models, respectively.
$\omega$ is used for weighting control.
To enhance the model’s sensitivity to different emotions under the same semantics, we design the negative visual prompt as an additional penalty term, computed as follows:
{
    \small
    \begin{equation}
    \begin{aligned}
        \widetilde{L}(\theta) & = \mathcal{L}(\theta) -\mathbb{E}_{\boldsymbol{x}_t^c,\boldsymbol{x}_t^r,c,t}[\log \sigma (- \beta T \omega ( \\ &
        (\lVert \boldsymbol{\epsilon}^c - \boldsymbol{\epsilon}_\theta(\boldsymbol{x}_t^c,c,t) \rVert_2^2 - \lVert \boldsymbol{\epsilon}^{n} - \boldsymbol{\epsilon}_\theta(\boldsymbol{x}_t^c,c_n,t) \rVert_2^2 ) ) ) ],
    \end{aligned}
    \label{eq:E-DPO}
    \end{equation}
}%
where $\boldsymbol{\epsilon}^{n}$ denotes the noise of the negative visual prompt.
In MC-Diffusion, the guidance condition $c$ is implemented using $p^t$ and $p^v$ as illustrated above.
The negative prompts $c_n$ utilize $p^t$ to provide the same semantics, and retrieve the most semantically related negative visual prompt $p^v_n$ from the bank of a randomly selected emotion $e_n$.
During training, the parameters of LoRA layers, $F^v_e$, linear layer mapping $F^v_e$ to $\mathcal{H}^v$ are optimized.
Note that the $F^t_e$ remains frozen, serving as the key for CLIP-encoded query retrieval.
\begin{figure*}[!t]
  \centering
  \includegraphics[width=0.92\linewidth]{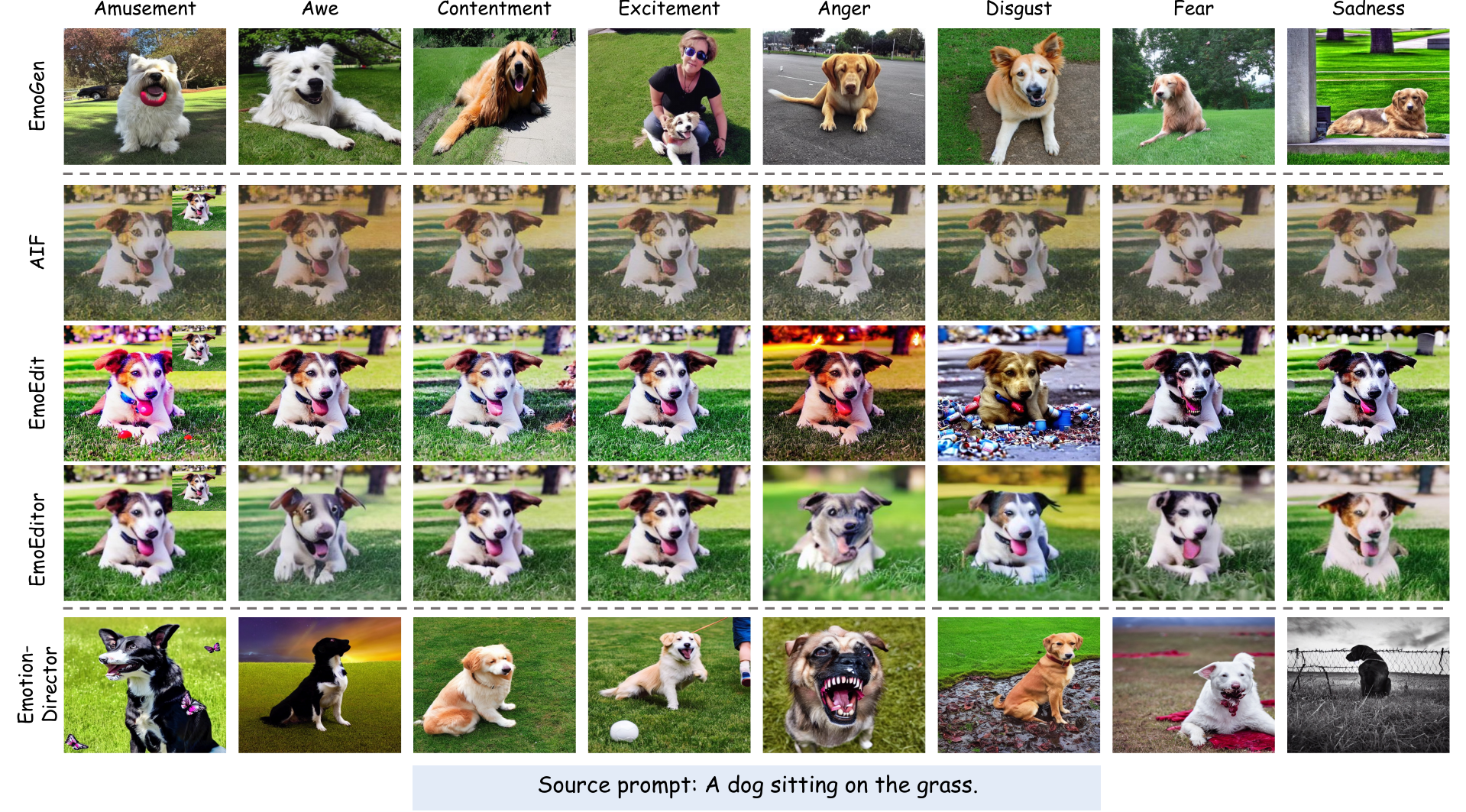}
   \vspace{-5pt}
   \caption{\small Qualitative comparison with emotion-oriented baselines.
   EmoGen aims to generate emotion-oriented images.
   AIF, EmoEdit, EmoEditor edit a source image into an emotion-oriented image, the source image is generated by SD v1.5.
   To ensure fairness and controllability, we follow EmoGen's emotion-transfer setting to generate images that match both the textual prompt’s semantics and the intended emotion.
   All methods use SD v1.5 for fairness, except AIF, which is non-diffusion.
   }
   \vspace{-15pt}
   \label{fig:emotion-comparison-1}
\end{figure*}

\subsection{MC-Agent}
\label{sec:MC-Agent}
MC-Agent builds a multi-agent system to address the expressive shortcut.
Most emotion-related expressions in the LLM rewritten prompts are template-like and difficult to visually represent, such as ``a pungent smell'' and ``a happy atmosphere''. 
To tackle this issue, we develop a multi-agent system to provide human-like multi-view reasoning capability, and design a chain-of-concept workflow, assisting LLM in generating emotion-highlighted and visually expressible textual prompts.
MC-Agent comprises four procedures: visual concept extraction, emotion attribution, textual prompt rewriting, and checking.
The source prompts are obtained from DiffusionDB~\cite{wang2023diffusiondb} by filtering according to NSFW scores and cosine similarity.
\noindent\textbf{Proc. 1: Visual concept extraction} aims to discover the visual concepts in the source prompts, \textit{such as humans, animals, and landscapes}.
%
% Visual concepts, such as humans, animals, and landscapes, serve as a bridge between textual descriptions and visual representations.
%
Visual concepts have been demonstrated to be crucial for conveying emotions in images~\cite{borth2013large}.
Therefore, we employ three agents to first extract the visual concepts in the source prompt as the cue for rewriting.
\noindent\textbf{Proc. 2: Emotion attribution} aims to provide human-like multi-view reasoning capability centered on visual concepts.
As is well known, emotion is subjective~\cite{zhao2016predicting}.
To avoid generating template-like rewritings, we randomly initialize three agents with different user backgrounds, \textit{e.g.}, \textit{age, gender, and profession}.
Each agent is instructed to provide a description based on its own background, analyzing the causes of the intended emotion evoked by the specified visual concept when reading the source prompt.
\noindent\textbf{Proc. 3: Textual prompt rewriting} aims to generate prompts that convey the intended emotion based on the given visual concept and candidate emotion attributions. 
Moreover, we explicitly instruct that the rewritten prompts must not conflict with or omit the semantics of the source prompts. 
\noindent\textbf{Proc. 4: Checking} aims to evaluate the quality of rewritten prompts, consisting of both textual and multi-modal feedback.
Textual feedback employs LLM-judge to evaluate the semantic consistency with source prompt, and the logical coherence of the rewritten prompt.
The emotion faithfulness of the rewritten prompt is evaluated by a model fine-tuned on the captions of affective images~\cite{achlioptas2021artemis, achlioptas2023affection}.
The multi-modal feedback evaluates the visual expressiveness of the rewritten prompt.
We use SDXL to generate four images for each rewriting prompt and compute the average CLIP score between prompts and images, filtering out prompts with weak visual relevance.
For each source prompt, we attempt up to three rounds of rewriting.
If no rewritten prompt passes the checking process after all attempts, the source prompt is discarded.
Moreover, considering the expensive cost of multi-agent for a single prompt, we employ MC-Agent to construct 80K high-quality rewritten textual prompts.
Then, we fine-tune a Qwen2.5-Instruct-72B model~\cite{qwen2.5} on the prompts to distill the rewriting capability into the model.
\begin{figure}[!t]
  \centering
  \includegraphics[width=1.0\linewidth]{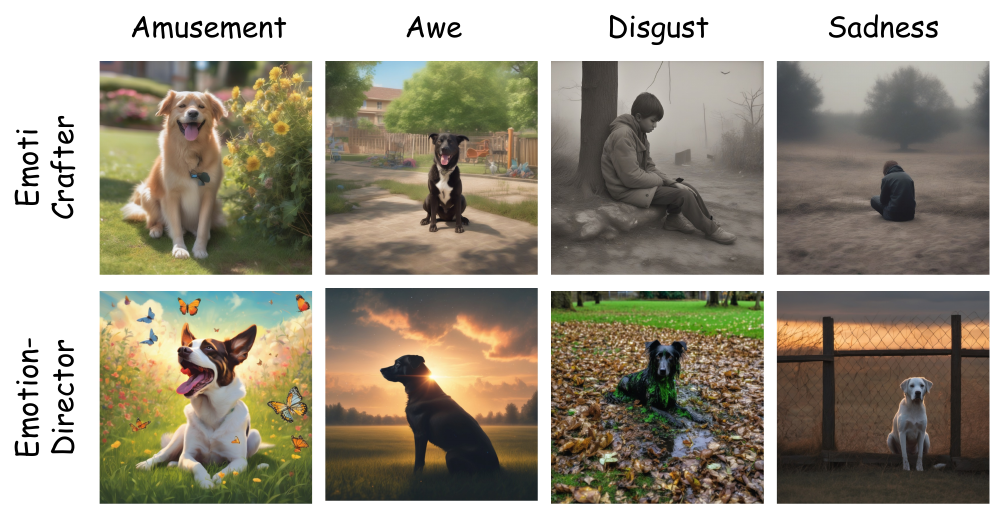}
   \vspace{-15pt}
   \caption{Qualitative comparison with EmotiCrafter. 
   Both methods use SDXL for fairness.
   }
   \vspace{-15pt}
   \label{fig:emotion-comparison-2}
\end{figure}

\section{Experiments}
\label{sec:experiment}
\subsection{Implementation Details}
%
%% 补充材料要加的细节：数据集的统计信息，比如每一类emotion有多少样本对、prompt也需要考虑一下怎么展示；多智能体中的LLM、微调的LLM、微调的学习率、优化器、调度器等；SD v1.5的结果;
%
For MC-Agent, we adopt Qwen2.5-Instruct-72B~\cite{qwen2.5} as the backbone LLM for agents, the fine-tuned model uses a LoRA rank of 8 for text rewriting.
We use Mikels model~\cite{mikels2005emotional}, which consists of eight emotions: amusement, awe, contentment, excitement, (\textit{four positive emotions}), anger, disgust, fear, and sadness (\textit{four negative emotions}).
MC-Diffusion adopts SD v1.5~\cite{rombach2022high} and SDXL~\cite{podellsdxl} as backbone, and leverage Adafactor~\cite{shazeer2018adafactor} as the optimizer~\cite{wallace2024diffusion}.
MC-Diffusion uses LoRA rank of 64, the same as SPO~\cite{liang2024step}.
All methods are fine-tuned on ETI for 2,000 steps.
The batch size is set to 64 for SFT, and 16 for all DPO-based methods.
The learning rate is set to 1e-5 with 25\% linear warmup. 
The KL regularization coefficient $\beta$ is set to 5000.
\noindent\textbf{Dataset}. 
Due to the lack of emotion-oriented text-to-image data and the preference pairs required for DPO fine-tuning, we construct an emotion-oriented text-to-image (ETI) dataset for training.
First, we select over 10K initial text prompts from DiffusionDB~\cite{wang2023diffusiondb} based on diversity and NSFW~\cite{kirstain2023pick}.
Then, MC-Agent rewrites each prompt into eight emotions, yielding 80K emotion-highlighted and visually expressible text prompts.
Next, we generate 16 candidate images per prompt using SD v1.5~\cite{rombach2022high}, SDXL~\cite{podellsdxl}, SD v3~\cite{esser2024scaling}, and FLUX~\cite{flux2024}, and annotate them with aesthetic scores~\cite{wallace2024diffusion}, text–image consistency~\cite{radford2021learning}, and human preference feedback~\cite{xu2023imagereward}.
For emotion annotation, we use a ViT-based emotion classifier and GPT-4o for coarse filtering, the retained images are then scored by 11 experts.
To obtain preference pairs, we select samples where the chosen image outperforms the rejected one on all metrics, and exceeds it by at least 0.2 in emotion score.
In total, we obtain 20,848 preference pairs, forming the ETI dataset used to train Emotion-Director.

\noindent\textbf{Baselines}. 
For a comprehensive evaluation, we compare two types of methods: emotion-oriented methods and training approaches.
The former consists of EmoGen~\cite{yang2024emogen}, EmotiCrafter~\cite{dang2025emoticrafter}, (\textit{two generation baselines}), AIF~\cite{weng2023affective}, EmoEdit~\cite{yang2025emoedit}, and EmoEditor~\cite{lin2024make}, (\textit{three editing baselines}).
The training algorithms include SFT, Diffusion-DPO~\cite{wallace2024diffusion}, and SPO~\cite{liang2024step}.

\begin{table}
  \centering
  \small
  \resizebox{0.49\textwidth}{!}{
    \begin{tabular}{lcccccc}
      \toprule
      \textbf{Method} & \textbf{HPSV2} & \textbf{IR} & \textbf{Sem-C} & \textbf{Sem-D} & \textbf{Emo-A} & \textbf{Emo-S}  \\
      \midrule
      \rowcolor[gray]{0.9}  \multicolumn{7}{l}{\textit{Affective image generation (SD v1.5)}} \\
      EmoGen & 22.79 & \underline{0.228} & 0.434 & \underline{0.036} & \underline{47.9} & \underline{63.5} \\
      \rowcolor[gray]{0.9}  \multicolumn{7}{l}{\textit{Affective image editing (SD v1.5)}} \\
      AIF & 23.03 & 0.105 & 0.413 & 0.016 & 20.9 & 37.2 \\
      EmoEdit & 23.26 & 0.206 & 0.425 & 0.022 & 41.4 & 49.7 \\
      EmoEditor & \underline{23.87} & 0.170 & \underline{0.442} & 0.026 & 34.7 & 46.5 \\      
      \midrule
      Emotion-Director & \textbf{24.65} & \textbf{0.318} & \textbf{0.475} & \textbf{0.044} & \textbf{54.1} & \textbf{66.1} \\
      \midrule
      \rowcolor[gray]{0.9}  \multicolumn{7}{l}{\textit{Affective image generation (SDXL)}} \\
      EmotiCrafter$^{\dagger}$ & 28.39 & 0.625 & 0.535 & 0.029 & 55.7 & 66.8 \\
      \rowcolor[gray]{0.9}  \multicolumn{7}{l}{\textit{Training algorithm (SDXL)}} \\
      SFT & 28.06 & 0.769 & \underline{0.559} & 0.044 & 49.8 & 61.8 \\
      Diffusion-DPO & 28.13 & \underline{0.882} & 0.543 & 0.040 & 52.3 & 64.5 \\
      SPO & \textbf{31.08} & 0.817 & 0.512 & 0.036 & 38.6 & 55.8 \\
      \midrule
      SFT$^{\dagger}$ & 27.91 & 0.714 & 0.537 & \underline{0.051} & 53.3 & 67.7 \\
      Diffusion-DPO$^{\dagger}$ & 28.07 & 0.859 & 0.540 & 0.047 & \underline{59.5} & \underline{70.4} \\
      SPO$^{\dagger}$ & \underline{30.29} & 0.801 & 0.514 & 0.035 & 40.2 & 62.4 \\
      \midrule
      Emotion-Director$^{\dagger}$ & 28.72 & \textbf{0.914} & \textbf{0.577} & \textbf{0.053} & \textbf{64.6} & \textbf{78.9} \\
      \bottomrule
    \end{tabular}
  }
  \vspace{-10pt}
  \captionof{table}{
      Quantitative evaluations.
      IR refers to ImageReward~\cite{xu2023imagereward}.
      Emo-S denote subjective evaluation of emotion score annotated by 11 experts.
      $\dagger$ means utilize our MC-Agent generated prompts.
  }
  \label{tab:comparison}
  \vspace{-20pt}
\end{table}

\begin{figure*}[!t]
  \centering
  \includegraphics[width=0.93\linewidth]{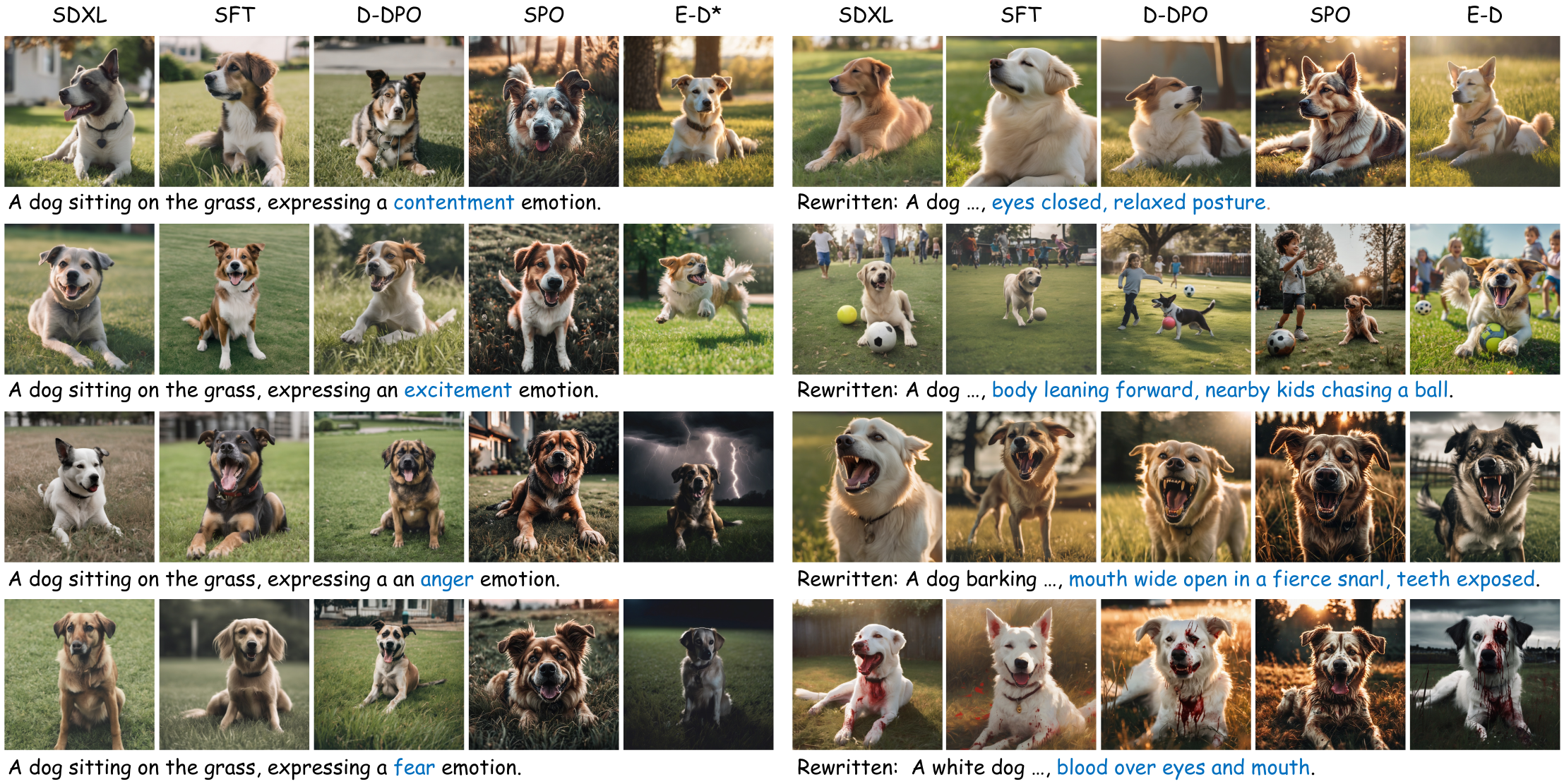}
   \vspace{-5pt}
   \caption{
       Qualitative comparison with three training algorithms, including SFT, diffusion-dpo (D-DPO), and SPO. E-D* denotes images generated by Emotion-Director without MC-Agent rewritten prompts.
   }
   \vspace{-15pt}
   \label{fig:comparison-2}
\end{figure*}
\subsection{Comparison with State-of-the-Art Methods}
For a comprehensive evaluation, we conduct both qualitative and quantitative experiments to compare Emotion-Director with state-of-the-art methods.
%
% The methods include SDXL~\cite{podellsdxl} (without fine-tuning), supervised fine-tuning (SFT), Diffusion-DPO~\cite{wallace2024diffusion}, and SPO~\cite{liang2024step}.
%
To ensure fairness, three training algorithms are fine-tuned on the ETI dataset, and all the methods are compared using the same backbone models.
Note that EmoGen, EmotiCrafter, and emotion-oriented editing methods require attribution annotation, valence-arousal annotation, and source image, respectively, and thus cannot be fine-tuned on the ETI dataset.
\noindent\textbf{Qualitative comparison}.
The initial 2,000 prompts for evaluation are selected from Pick-a-Pic~\cite{kirstain2023pick}.
According to the qualitative comparisons shown in Fig.~\ref{fig:emotion-comparison-1}, \ref{fig:emotion-comparison-2}, and \ref{fig:comparison-2}, we can observe that
1) \textit{\textbf{Compared with emotion-oriented methods, Emotion-Director shows clear advantages in both emotion expression and instruction following.}}
Image-editing approaches adjust colors and local details, which fail in most cases.
Since ``dog'' is included in the aligned attributes of amusement and contentment, EmoGen achieves higher faithfulness on these emotions, but it still cannot generate others such as awe and anger.
EmotiCrafter uses SDXL as its backbone, yielding noticeably improved image quality.
However, its emotion faithfulness remains insufficient, and its instruction-following ability degrades substantially (\textit{e.g.}, disgust, sadness).
2) \textit{\textbf{Without MC-Agent rewritten prompts, Emotion-Director can still express the intended emotion.}}
Supported by the visual prompts, Emotion-Director adjusts image brightness and incorporates typical visual elements commonly associated with the intended emotion (\textit{e.g.}, lightning for anger) to generate emotion-highlighted images.
3) \textit{\textbf{With MC-Agent–rewritten prompts, Emotion-Director still shows clear advantages}}.
Although the rewritten prompts provide the semantics needed for expressing emotion, the lack of visual prompts causes the lighting and atmosphere of the generated images that do not match the intended emotion.
By combining MC-Diffusion with MC-Agent, Emotion-Director improves both the semantics and the visual rendering required for affective expression, thereby producing emotion-highlighted images.

\noindent\textbf{Quantitative comparison}.
The metrics for quantitative comparison include commonly used~\cite{wallace2024diffusion, liang2024step} human feedback metrics HPSv2~\cite{wu2023human} and ImageReward~\cite{xu2023imagereward}, as well as three emotion faithfulness metrics Sem-C, Sem-D, and Emo-A~\cite{yang2024emogen, yang2025emoedit}.
Besides, considering the human-centered characteristic of emotion, we employ 11 experts for subjective evaluation.
According to the results shown in Tab.~\ref{tab:comparison}, we observe that:

\begin{table}
  \centering
  \small
  % \vspace{-5pt}
  \resizebox{0.49\textwidth}{!}{
    \begin{tabular}{lcccccc}
      \toprule
      \textbf{Setting} & \textbf{HPSV2} & \textbf{IR} & \textbf{Sem-C} & \textbf{Sem-D} & \textbf{Emo-A} \\
      \midrule
      Emotion-Director & 28.72 & 0.914 & 0.577 & 0.053 & 64.6  \\
      \rowcolor[gray]{0.9}  \multicolumn{6}{l}{\textit{Module-level ablation}} \\
      w/o MC-Diffusion & 27.18 & 0.672 & 0.522 & 0.051 & 55.8 \\
      w/o MC-Agent & 28.05 & 0.840 & 0.569 & 0.045 & 58.9 \\
      SDXL & 27.91 & 0.635 & 0.515 & 0.040 & 46.1 \\
      \rowcolor[gray]{0.9}  \multicolumn{6}{l}{\textit{MC-Diffusion ablation}} \\
      w/o v-prompt & 28.07 & 0.859 & 0.540 & 0.047 & 59.5 \\
      w/o DPO & 27.62 & 0.736 & 0.564 & 0.053 & 60.4 \\
      w/o neg. & 28.54 & 0.844 & 0.577 & 0.051 & 62.6 \\    
      \rowcolor[gray]{0.9}  \multicolumn{6}{l}{\textit{MC-Agent ablation}} \\
      w/o Agents & 28.35 & 0.861 & 0.579 & 0.039 & 59.5 \\
      w/o CoC & 28.09 & 0.773 & 0.564 & 0.049 & 61.2 \\
      \bottomrule
    \end{tabular}
  }
  \vspace{-6pt}
  \captionof{table}{
      Ablation of our Emotion-Director.
      v-prompt and neg. denote the visual and negative visual prompts in the improved DPO loss.
      w/o DPO applies SFT for training.
      %
      % Removing visual prompts falls back to Diffusion-DPO.
      %
      w/o Agents uses a single LLM for rewriting. 
      CoC denotes the chain-of-concept, removing it skips visual-concept extraction and concept-related instructions.
  }
  \vspace{-15pt}
  \label{tab:ablation}
\end{table}

\noindent 1) \textit{\textbf{Emotion-Director achieves comparable performance on human-feedback metrics that are not emotion-related}}. Notably, SPO substantially improves image aesthetics and achieves the best results on HPSv2.
2) \textit{\textbf{Compared with emotion-oriented methods, Emotion-Director achieves the best performance on emotion-related metrics}}. 
Image editing methods generate highly similar images across different emotions, leading to a substantial drop in semantic diversity (Sem-D).
Moreover, our method yields a remarkable improvement on Emo-A, achieving at least a 6.2\% gain.
3) \textit{\textbf{Compared with other training algorithms, Emotion-Director shows clear superiority on emotion-related metrics}}.
Under the fair setting, where other methods do not use MC-Agent–rewritten prompts, our method achieves over 10\% improvements on both Emo-A and Emo-S.
Even when other methods use rewritten prompts, benefiting from visual prompts and the improved DPO loss, Emotion-Director still achieves at least a 5.1\% gain.

\begin{table}
\renewcommand{\arraystretch}{1.01}
  \centering
  \resizebox{0.49\textwidth}{!}{
    \begin{tabular}{lcccccc}
      \toprule
      \textbf{Method} & \textbf{Size} & \textbf{Emotion}~$\uparrow$ & \textbf{Pass$_{Con.}$}~$\uparrow$ & \textbf{Abs.}~$\downarrow$  & \textbf{Sim.}~$\downarrow$ & \textbf{Pass$_{Len.}$}~$\uparrow$ \\
      \midrule
      Qwen2.5 & 72B & 78.13 & 95.9 & 0.710 & 0.622 & \underline{99.63} \\
      D-Qwen2.5 & 32B & 71.63 & 93.6 & 0.714 & 0.565 & 91.25 \\
      GPT-4o & - & 82.25 & \textbf{98.1} & 0.709 & 0.580 & \textbf{100.0} \\
      CAMEL & - & 83.63 & 97.3 & 0.696 & 0.593 & 92.50 \\
      \midrule
      Qwen2.5$^{\ddagger}$ & 72B & \underline{85.65} & 96.5 & \underline{0.651} & \underline{0.576} & \textbf{100.0} \\
      MC-Agent & 72B & \textbf{87.75} & \underline{97.7} & \textbf{0.633} & \textbf{0.541} & \textbf{100.0} \\
      \bottomrule
    \end{tabular}
    }
    \vspace{-5pt}
    % \captionsetup{font=scriptsize}
    \caption{
    % \scriptsize
      Ablation on textual prompt rewriting. 
      Pass$_{Con.}$ and Pass$_{Len.}$ measure the pass rate of prompts that do not conflict with the original semantics, and the length ($<$70) rate to avoid truncation.
      Pass$_{Con.}$ is averaged over Qwen2.5 and GPT-4o to mitigate model bias.
      Abs. denote the abstractness (visual expressiveness), which is measured by abstractness lexicon~\cite{achlioptas2021artemis}.
      Sim. indicates similarity of rewritten prompts, where lower is more diverse.
    }
    \vspace{-15pt}
  \label{tab:text_ablation}
\end{table}

\subsection{Ablation Study}
We conduct both qualitative and quantitative experiments to evaluate the contribution of each component in our Emotion-Director.
According to the results shown in Tab.~\ref{tab:ablation} and Fig.~\ref{fig:ablation}, we can observe:
\begin{figure*}[!t]
  \centering
  \includegraphics[width=0.96\linewidth]{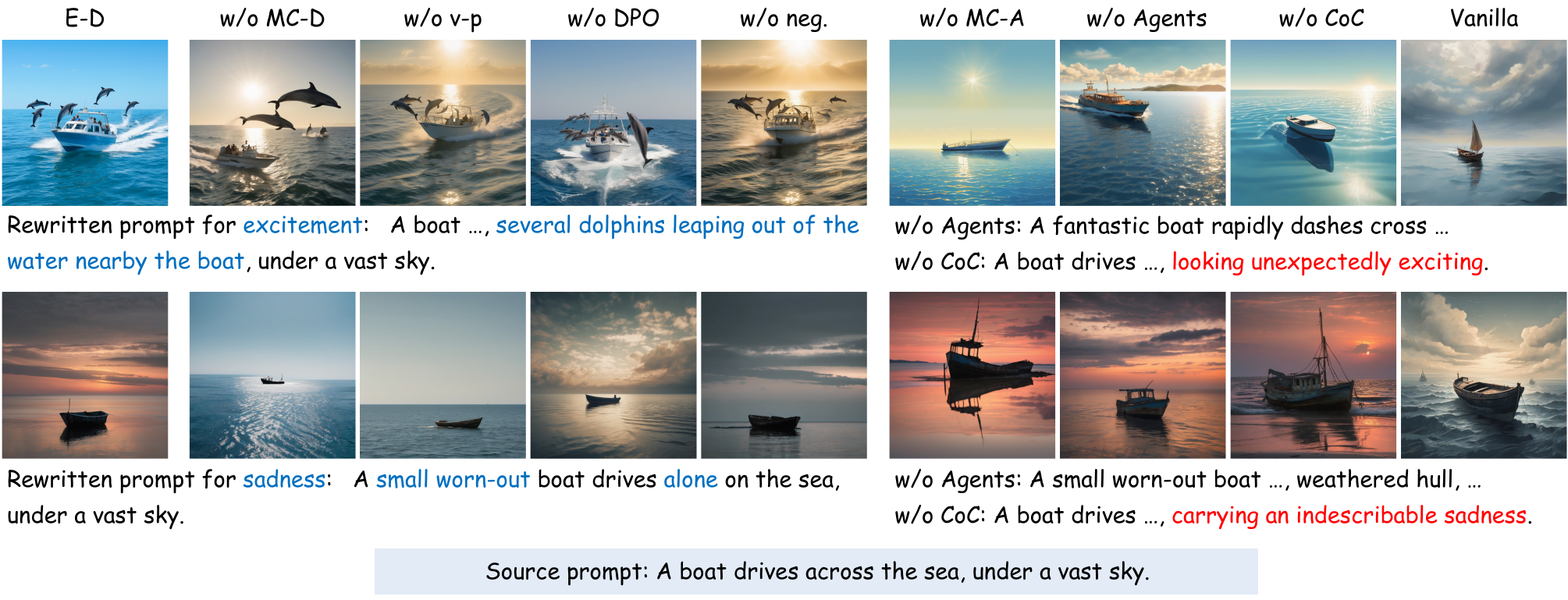}
   \vspace{-10pt}
   \caption{Qualitative ablation of our Emotion-Director.
   % 
   % Note that to ensure fairness, the results of w./o. revision are obtained from the models fine-tuned on ETI using DPO.
   %
   % TPR enhances emotional expression but may cause visual inconsistencies, while MMP effectively improves visual coherence, and combining both significantly boosts emotion-driven image generation.
   }
   \vspace{-10pt}
   \label{fig:ablation}
\end{figure*}
\noindent 1) \textit{\textbf{At module level, MC-Diffusion has a substantial impact on ImageReward and Sem-C, and MC-Agent clearly affects Sem-D}}.
MC-Diffusion improves text–image alignment (ImageReward) and semantic clarity (Sem-C) by fine-grained alignment through DPO fine-tuning.
In contrast, MC-Agent enriches the semantic content by rewriting prompts, thereby increasing semantic diversity (Sem-D).
2) \textit{\textbf{At component level, the visual prompt has a remarkable effect on MC-Diffusion’s capability to express emotions}}.
Without the visual prompt, the negative visual term in the improved DPO cannot be used, causing the model to fall back to Diffusion-DPO. 
In this case, the brightness of the generated images clearly fails to match the intended emotion, and the Emo-A metric drops noticeably.
3) \textit{\textbf{At component level, both the multi-agent system and the chain-of-concept workflow are essential}}.
Without the multi-agent system (\textit{i.e., using a single LLM}), the model tends to focus on a single subject, leading to a clear drop in the diversity of rewritten prompts. 
Without the chain-of-concept workflow, the rewritten prompts become overly abstract and difficult to express visually.

\begin{figure}[t!]
    \centering
    \includegraphics[width=0.93\linewidth]{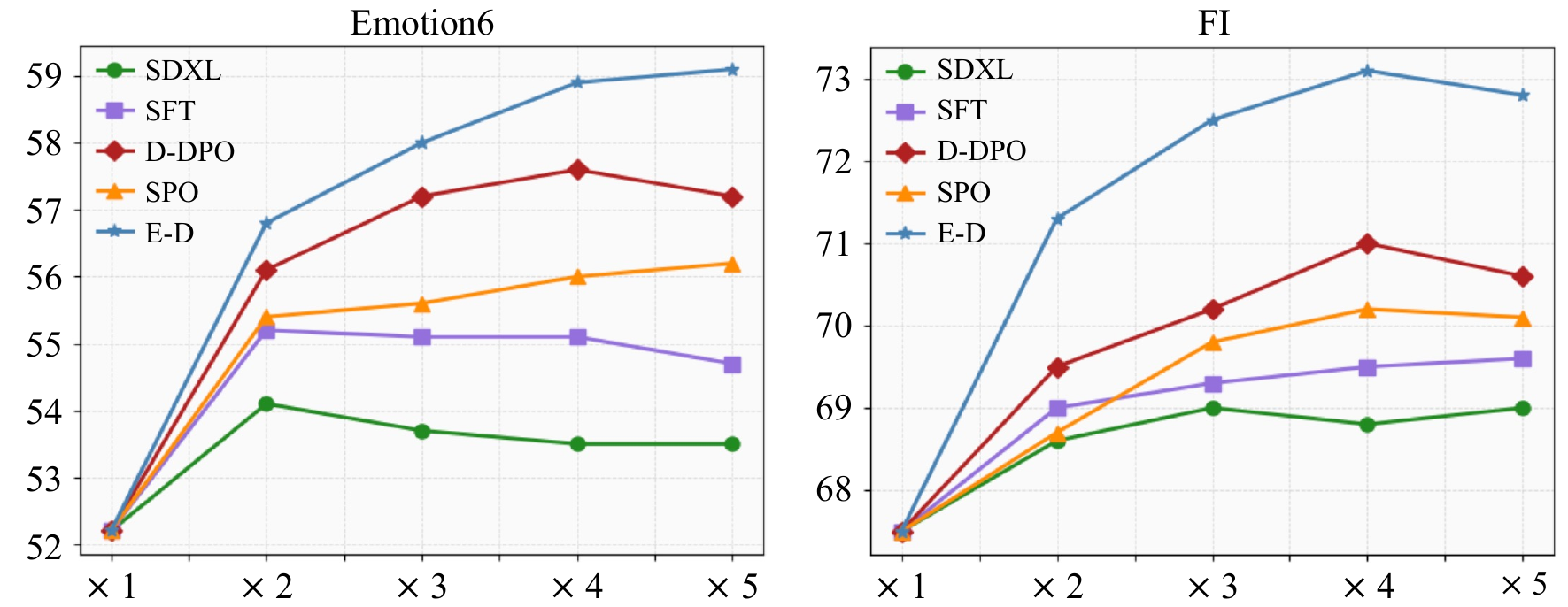}
    \vspace{-6pt}
    % \captionsetup{font=scriptsize}
    \caption{Application in image emotion recognition.
    $\times N$ refers to expanding the dataset by a factor of $N$.
    The synthesized data are generated from captions of original images by GPT-4o.
    }
    \label{fig:application}
    \vspace{-15pt}
\end{figure}

Besides, we conduct ablations on the prompt-rewriting model.
According to Tab.~\ref{tab:text_ablation}, we observe that:
1) With the chain-of-concept workflow, \textit{\textbf{MC-Agent yields a clear reduction in abstractness}}.
2) With the multi-agent system, our method achieves the best performance in semantic diversity and emotion metrics.

\subsection{Application}
We explore the applicability of emotion-oriented image generation to the emotion recognition task, the results are presented in Fig.~\ref{fig:application}.
Experiments are conducted on a small-scale Emotion6 dataset (1,980) and a widely used FI dataset (23,308).
Experimental results demonstrate that utilizing synthetic images is beneficial for emotion recognition, and leads to a higher gain on the small-scale dataset.
Among all methods, Emotion-Director yields the highest improvement, attributed to its powerful capacity for emotional expression.

\section{Conclusion}
\label{sec:conclusion}
In this paper, we explore the affective shortcut in previous emotion-oriented image generation methods, and propose Emotion-Director to generate emotion-highlighted images.
The Emotion-Director consists of MC-Diffusion and MC-Agent.
MC-Diffusion integrates visual prompts into guidance to generate affective images beyond semantics, and utilizes improved DPO loss to enhance the model's sensitivity to different emotions under the same semantics.
MC-Agent utilizes a chain-of-concept multi-agent workflow to generate emotion-highlighted and visually expressive prompts.
In the future, we will explore emotion-oriented video generation, which presents a substantial challenge in the temporal variation of emotion.
%
% Emotion-Director provides a foundational resource and an effective methodology, which is promising for emotion-oriented visual generation.
%

{
    \small
    \bibliographystyle{ieeenat_fullname}
    \bibliography{main}
}


\end{document}